\pdfoutput=1

\documentclass[11pt]{article}

\usepackage[]{acl}

\usepackage{times}
\usepackage{latexsym}
\usepackage{microtype}
\usepackage{graphicx}
\usepackage{subfigure}
\usepackage{amsmath}
\usepackage{amsthm}
\usepackage{amsfonts}
\usepackage{stfloats}
\usepackage{multirow}
\usepackage{float}
\usepackage{enumitem}
\usepackage{booktabs} 
\usepackage{tikz}
\usepackage{dashbox}
\usepackage{hyperref}
\usepackage[colorinlistoftodos,prependcaption,textsize=tiny,disable]{todonotes}

\usepackage[T1]{fontenc}

\usepackage[utf8]{inputenc}

\usepackage{microtype}

\definecolor{classes}{HTML}{1D741B}
\definecolor{arrow}{HTML}{86888C}
\definecolor{context}{HTML}{004369}
\definecolor{attributes}{HTML}{98042D}
\definecolor{type}{HTML}{98042D}
\usepackage{xcolor}

\usepackage{pifont}
\newcommand{\cmark}{\ding{51}}%
\newcommand{\xmark}{\ding{55}}%

\newcommand{\xhdr}[1]{\noindent{\bfseries #1}.}
\newcommand{\xhdrn}[1]{\noindent{\bfseries #1}}

\newcommand{\name}{REMix}
\newcommand{\smallsize}{\fontsize{10pt}{11pt}\selectfont}

\title{Semi-supervised Relation Extraction via \\
Data Augmentation and Consistency-training}

\author{Komal K. Teru \\
  The Vanguard Group \\
  \texttt{komal\_teru@vanguard.com} \\}

\begin{document}
\maketitle
\begin{abstract}
Due to the semantic complexity of the Relation extraction (RE) task, obtaining high-quality human labelled data is an expensive and noisy process.
To improve the sample efficiency of the models, semi-supervised learning (SSL) methods aim to leverage unlabelled data in addition to learning from limited labelled data points.
Recently, strong data augmentation combined with consistency-based semi-supervised learning methods have advanced the state of the art in several SSL tasks. However, adapting these methods to the RE task has been challenging due to the difficulty of data augmentation for RE.
In this work, we leverage the recent advances in controlled text generation to perform high quality data augmentation for the RE task. 
We further introduce small but significant changes to model architecture that allows for generation of more training data by interpolating different data points in their latent space. These data augmentations along with consistency training result in very competitive results for semi-supervised relation extraction on four benchmark datasets.


\end{abstract}

\section{Introduction}
Relation extraction is one of the essential components in constructing structured knowledge bases \cite{luan-etal-2018-multi}, performing interpretable question answering \cite{pmlr-v139-sun21e}, improving web search, and many other information extraction pipelines.
It aims to discover the semantic relation between a given head entity and tail entity based on the context in the input sentence. For example, given a sentence "\textit{The \underline{battle} led to \underline{panic} on the frontier, and settlers in the surrounding counties fled.}", the goal is to extract the \texttt{Cause-Effect} relation between the head entity `\textbf{battle}' and the tail entity `\textbf{panic}'.
The RE task requires a high level of language understanding and involves a significant level of semantic complexity \cite{re_review}.
Due to this semantic complexity it often requires extensive and highly skilled human involvement to obtain good quality labelled data, making data collection an expensive and noisy process.
Unsurprisingly, because of the same semantic complexity of the task, models typically require large amounts of labelled data to give production-ready performance.

\definecolor{bronze}{HTML}{E48400}
\begin{figure}[t!]
\fbox{
\begin{minipage}[b]{0.95\linewidth}
\begin{flushleft}
\smallsize \textcolor{context}{\textbf{Input sentence:}} The \textcolor{type}{\textbf{battle}} led to \textcolor{type}{\textbf{panic}} on the frontier, and settlers in the surrounding counties fled.
\end{flushleft}

\begin{flushleft}
\smallsize \textcolor{context}{\textbf{Synonym-replacement:}} The struggle cause scare on the frontier, and settlers in the surrounding counties fly.
\end{flushleft}

\begin{flushleft}
\smallsize \textcolor{context}{\textbf{LM-based augmentation:}} The \textcolor{type}{\textbf{battle}} reduced \textcolor{type}{\textbf{panic}} on the frontier, and settlers in the surrounding counties relaxed.
\end{flushleft}

\begin{flushleft}
\smallsize \textcolor{context}{\textbf{Vanilla BT:}} The war caused \textcolor{type}{\textbf{panic}} at the border, and residents of the nearby counties fled.
\end{flushleft}
\hrule
\begin{flushleft}
\smallsize \textcolor{context}{\textbf{Our constrained BT:}} The \textcolor{type}{\textbf{battle}} sparked \textcolor{type}{\textbf{panic}} at the border, with residents fleeing in surrounding counties.
\end{flushleft}
\end{minipage}
}
\caption{Different data augmentation techniques applied to a sample datapoint from SemEval dataset. Existing methods replace the head/tail entities (highlited in red), change the original meaning or do not give very fluent paraphrases.}
\label{fig:re_aug}
\end{figure}

A common strategy to improve the sample efficiency of machine learning models is semi-supervised learning methods which leverage easily accessible unlabelled data to improve the overall performance.
While there are several paradigms of semi-supervised learning methods, consistency training based methods have advanced the state of the art in several SSL tasks \cite{ghosh2021on}. These methods can typically reach performances that are comparable to their fully supervised counterparts while using only a fraction of labelled data points.
Recently, strong data augmentation combined with consistency training algorithms have shown great success, even surpassing fully supervised models, in low-data settings of various tasks \cite{uda}. 
Adapting these methods to the task of relation extraction has been challenging due to the difficulty of data augmentation for RE task.
This is because, in addition to the input sentence, each data point also consists of a head entity and a tail entity contained in the input sentence. Typical data augmentation techniques used in NLP such as back-translation, synonym-replacement, language-model based augmentation, etc. \cite{feng-etal-2021-survey} can not be easily applied to such `structured' input as they do not guarantee the integrity of either a) the entities in the input sentence or, b) the meaning of the input sentence itself. 
Figure \ref{fig:re_aug} shows that using synonym-replacement and vanilla back-translation (BT) methods \cite{nlpaug-bt} the entities themselves could be paraphrased or replaced. Matching the new and the old entities is a whole problem in itself. In the Language Model-based augmentation method \cite{lambada}, the semantic meaning of the input sentence changes altogether, which makes it difficult to employ consistency training.

\xhdr{Present work} In this work, we leverage the recent advances in controlled text generation to perform high quality data augmentation for the relation extraction task that not only keeps the meaning and the head/tail entities intact but also produces fluent and diverse data points.
In particular, we modify back-translation to leverage lexically constrained decoding strategies \cite{constrained_decoding_1, constrained_decoding_2} in order to obtain paraphrased sentences while retaining the head and the tail entities. 
We further propose novel modifications to the widely popular relation extraction model architecture, that allows for generation of more samples by interpolating different data points in their latent space, a trick that has been very successful in other domains and tasks \cite{mixmatch, mixtext, lada}. 
Additionally, we leverage the entity types of the head and the tail entities, when available, in a way that effectively exploits the knowledge embedded in pre-trained language models.
These data augmentations, when applied to unlabelled data, let us employ consistency training techniques to achieve very competitive results for semi-supervised relation extraction on four benchmark datasets.
To the best of our knowledge, this is the first study to apply and show the merit of data augmentation and consistency training for semi-supervised relation extraction task.


\section{Related work}
\subsection*{Semi-supervised learning for NLP}
Semi-supervised learning algorithms can be categorized into two broad classes--1) \textit{self-ensembling}  methods and 2) \textit{self-training} methods.

\xhdrn{\textit{Self-ensembling} methods} leverage the smoothness and cluster/low-density assumptions of the latent space \cite{pmlr-vR5-chapelle05b}.
They train the models to make consistent predictions under various kinds of perturbations to either a) the data \cite{vat, uda}, or b) the model parameters themselves \cite{mt}. The former methods are broadly referred to as consistency training methods and have resulted in state-of-the-art performances for several semi-supervised NLP tasks.
\citet{Sachan2019RevisitingLN} add adversarial noise to both labelled and unlabelled data and train models to make consistent predictions on the original and the corresponding noisy data-point. 
Many recent methods leverage large pre-trained language models for more advanced data augmentation techniques, like back-translation \cite{Edunov2018UnderstandingBA}, and further improve performance in the low-data regime \cite{uda}.
Recently, \citet{mixtext, lada} adapted the Mixup algorithm \cite{mixup} as another form of data augmentation for textual data and show state-of-the-art performance on text classification and NER tasks.
Due to the difficulty of data augmentation for relation extraction task (Figure \ref{fig:re_aug}), these methods have not been adapted for semi-supervised relation extraction (SSRE) task so far. In this work, we fill that gap and demonstrate the empirical success of consistency training for SSRE.

\xhdrn{\textit{Self-training} methods} are the oldest heuristic methods of iteratively expanding the labelled training set by including high-confidence \textit{pseudo-labels} from the unlabelled data.
All of the existing works on SSRE fall under this paradigm. 
These methods famously suffer from the confirmation bias problem where the incorrect predictions of the initially trained model affect the quality of pseudo-labels and eventually cause the label distribution to drift away from the true data distribution, resulting in a \textit{semantic drift}. \citet{DualRE} was one of the first works to address this by training two different models and augmenting the labelled set with the `consensus' set, i.e., the data points which are labelled the same by both models. Several works have developed on this idea of improving the pseudo-label quality via various strategies like meta-learning \cite{Hu2021SemisupervisedRE} or reinforcement learning \cite{Hu2021GradientIR}. 
These set of methods constitute our baselines.


\subsection*{Data augmentation for NLP}
In this work, we concentrate on two major classes of data augmentation techniques for NLP -- sentence-level data augmentation and latent space augmentations. Sentence-level data augmentation techniques include back-translation \cite{Edunov2018UnderstandingBA}, language-model based augmentations \cite{lambada}, and word-replacement strategies \cite{nlpaug-syn-rep}. We adapt the back-translation techniques to the RE task.

In latent space augmentations one generates more samples by interpolating between pairs of given data points in their latent space.
This was originally introduced for image classification \cite{mixup, manifold_mixup, cutmix} as a data augmentation and regularization method.
Previous works have generalized this idea to the textual domain by proposing to interpolate in embedding space \cite{advaug} or the general latent space \cite{mixtext, lada} of textual data and applied the technique to NLP tasks such as text classification, machine translation, NER task and achieved significant improvements. We show that both these styles of augmentations can be effectively applied to improve performance on SSRE task.

\section{Background}
\label{sec:base_model}

\xhdr{Task formulation} In this work, we focus on the sentence-level relation extraction task, i.e., given a \textit{relation statement} $\mathbf{x}: (\mathbf{s}, e_h, e_t)$ consisting of a sentence, $\mathbf{s}$, a head entity, $e_h$, and a tail entity, $e_t$ (both the entities are mentioned in the given sentence $\mathbf{s}$), the goal is to predict a relation $r \in \mathcal{R} \cup \{\text{NA}\}$ between the head and the tail entity, where $\mathcal{R}$ is a pre-defined set of relations. If the sentence does not express any relation from the set $\mathcal{R}$ between the two entities, then the relation statement $\mathbf{x}$ is accordingly labelled NA. 

This is typically done by learning a relation encoder model $\mathcal{F}_{\theta}: \mathbf{x} \mapsto \mathbf{h}_r$ that maps an input relation statement, $\mathbf{x}$, to a fixed length vector $\mathbf{h}_r$ that represents the relation expressed in $\mathbf{s}$ between $e_h$ and $e_t$. This relation representation, $\mathbf{h}_r$, is then classified to a relation $r \in \mathcal{R} \cup \{\text{NA}\}$ via an MLP classifier.

\xhdr{Base model architecture} Most recent methods for RE use a Transformer-based architecture \cite{bert, vaswani2017attention} for the relation encoder model, $\mathcal{F}_{\theta}$.
To represent the head and tail entities in the input to the encoder, the widely accepted strategy is to augment the input sentence $\mathbf{s}$ with entity marker tokens--\texttt{[E1],} \texttt{[/E1]}, \texttt{[E2],} \texttt{[/E2]}--to mark the start and end of both entities.
Concretely, an input sentence like "\textit{\underline{Lebron James} currently plays for \underline{LA Lakers} team.}" when augmented with entity marker tokens becomes
\begin{multline*}
\text{\texttt{[E1]} \textbf{Lebron James} \texttt{[/E1]} currently} \\
\text{plays for \texttt{[E2]} \textbf{LA Lakers} \texttt{[/E2]} team.}
\end{multline*}
This modified text is input to the Transformer-based sequence encoder.
Next, the encoder output representations\footnote{hidden state from the last layer of the Transformer model} of the tokens \texttt{[E1]} and \texttt{[E2]} are concatenated to give the fixed length relation representation, $\mathbf{h}_r = [\mathbf{h}_{[E1]} \oplus \mathbf{h}_{[E2]}]$. This fixed length vector is in turn passed through an MLP classifier, $p_\phi(\mathbf{h}_r)$, to give a probability vector, $\mathbf{y}$, over the relation set $\mathcal{R} \cup \{\text{NA}\}$.

\section{Proposed approach}

In our approach we build on the base model architecture described in \S\ref{sec:base_model} and introduce additional model design elements that are necessary to obtain an improved performance in semi-supervised relation extraction (SSRE) task.

We first describe the two data augmentation techniques we perform, and the model architectural changes we introduce that facilitate these augmentations. 
Then, we describe the training procedure we follow to leverage unlabelled data and achieve state-of-the-art performance on three out of four benchmark datasets for SSRE.

\subsection{Constrained back-translation}
\begin{figure}[t]
  \centering
  \centerline{\includegraphics[width=\linewidth]{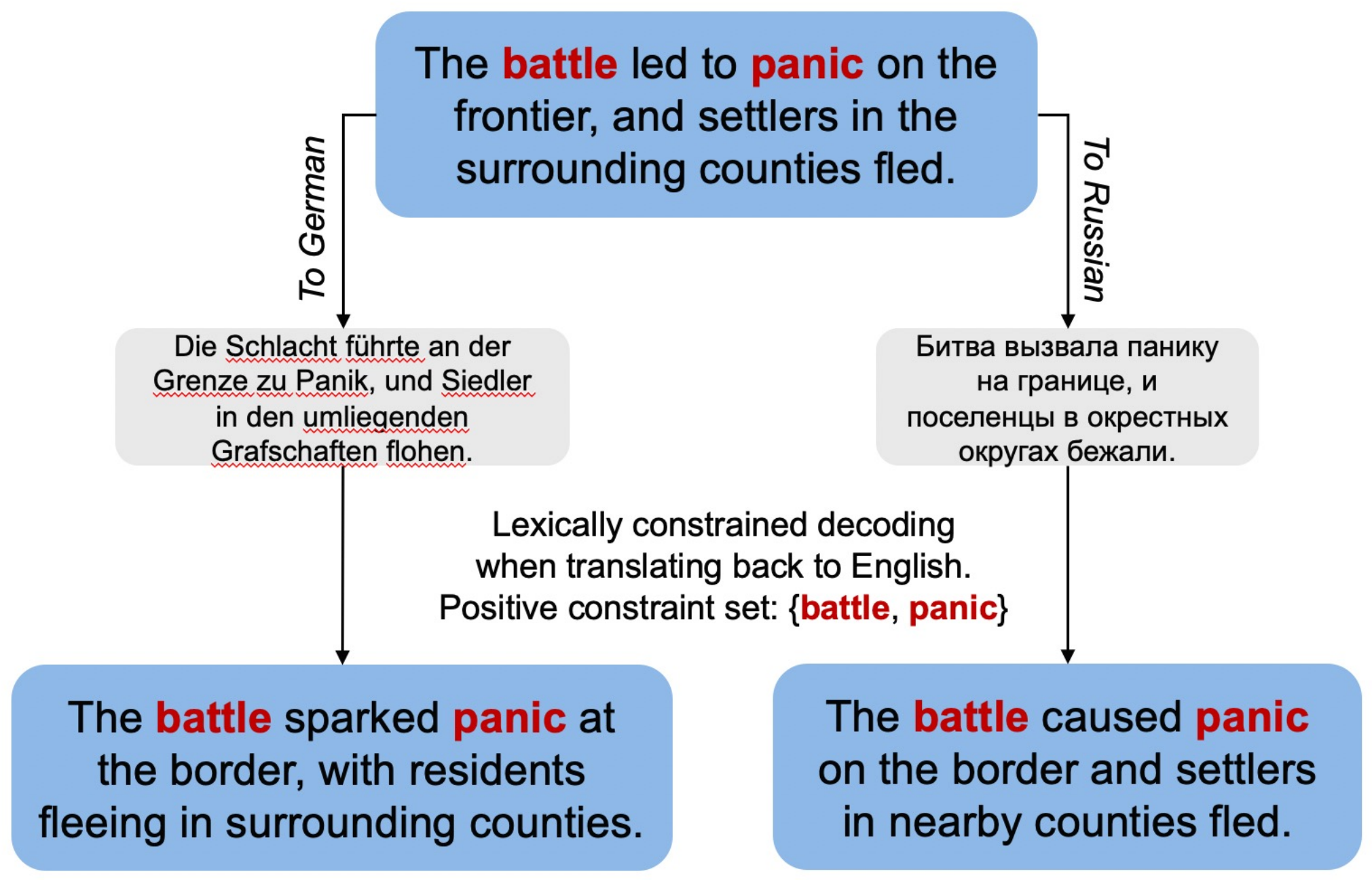}}
\caption{Constrained back-translation process}
\label{fig:btda}
\end{figure}

Back-translation \cite{Edunov2018UnderstandingBA} generates diverse and fluent augmentations while retaining the global semantics of the original input.
Specifically, one translates a given text into an intermediate language, say, German, and translates it back to the source language, say English.
Using different intermediate languages and temperature-based sampling results in a diverse set of paraphrases.
Applying this back-translation technique in a vanilla fashion is not possible for RE task because one has little control over the retention of the head and tail entities (Figure \ref{fig:re_aug}). Thus, when translating back to the source language from the intermediate language we perform lexically-constrained decoding \cite{constrained_decoding_2}, i.e., force the inclusion of pre-specified words and phrases--positive constraint set--in the output. In our case the original head and tail entity words/phrases make up this positive constraints set. 
We use German and Russian as intermediate languages and use the pre-trained WMT’19 English-German and English-Russian translation models (in both directions) and their implementations provided by \citet{ott2019fairseq}. \todo{Could be moved to footnote}
This methodology generates diverse data augmentations for a given sentence. For example, the sentence "\textit{The battle led to panic on the frontier, and settlers in the surrounding counties fled.}" is converted to "\textit{The battle sparked panic at the border, with residents fleeing in surrounding counties}" when back-translated via German, and to "\textit{The battle caused panic on the border and settlers in nearby counties fled.}" when done via Russian. This is illustrated in Figure \ref{fig:btda}.
This strong data-augmentation technique for RE can be applied to both labelled and unlabelled data opening the doors to consistency training \cite{uda} as we will see in \S\ref{sec:ssre}.

\subsection{Latent-space interpolation}
\label{sec:mixre}
Here, we adapt a mixup-based data augmentation technique to the RE task by making necessary modifications to the base model architecture we described in \S\ref{sec:base_model}. 
As done in previous works \cite{mixtext, lada}, we sample two random data points--$(\mathbf{x}, \mathbf{y})$ and $(\mathbf{x}^\prime, \mathbf{y}^\prime)$, where $\mathbf{x}$ and $\mathbf{y}$ denote the relation statement and the corresponding relation label--from the training data and separately compute the respective latent representations, $\mathbf{h}^m$ and $\mathbf{h}^{\prime m}$, upto the layer $m$ of the relation encoder $\mathcal{F}_{\theta}$ as follows:
\begin{align*}
    \mathbf{h}^l &= \mathcal{F}_{\theta}^l(\mathbf{h}^{l-1}); \quad l \in [1, m], \\
    \mathbf{h}^{\prime l} &= \mathcal{F}_{\theta}^l(\mathbf{h}^{\prime {l-1}}); \quad l \in [1, m],
\end{align*}
where $\mathbf{h}^l$ is the latent representation of all tokens in the sentence $\mathbf{x}$ at the $l^\text{th}$ layer of the encoder.
Next, the latent representations of each token in $\mathbf{x}$ at the $m^{\text{th}}$ layer are linearly interpolated:
$$
\tilde{\mathbf{h}}^m = \lambda\mathbf{h}^m + (1 - \lambda)\mathbf{h}^{\prime m},
$$
where $\lambda$ is the mixing coefficient which is sampled from a Beta distribution, i.e., $\lambda \sim \text{Beta}(\alpha, \beta)$. Then, the interpolated latent representation is passed through the rest of the encoder layers:
$$
\tilde{\mathbf{h}}^l = \mathcal{F}_{\theta}^l(\tilde{\mathbf{h}}^{l-1}); \quad l \in [m + 1, L].
$$
This final encoder output representation, $\tilde{\mathbf{h}}^L$, can be interpreted as the encoder output representation of a \textit{virtual} input $\tilde{\mathbf{x}}$, i.e., $\tilde{\mathbf{h}}^L = \mathcal{F}_{\theta}(\tilde{\mathbf{x}})$.
We denote this whole mixup operation\footnote{\name\ entails the model architecture changes discussed below.} as $\tilde{\mathbf{h}}^L := \text{\name}(\mathbf{x}, \mathbf{x}^\prime)$.
The label for this augmented \textit{virtual} sample is given by the linear interpolation of the respective labels, $\mathbf{y}$ and $\mathbf{y}^\prime$, with the same mixing coefficient $\lambda$ i.e., $\tilde{\mathbf{y}} := \text{mix}(\mathbf{y}, \mathbf{y}^\prime) = \lambda \mathbf{y} + (1 - \lambda) \mathbf{y}^\prime$.
This \textit{virtual} data point, $(\tilde{\mathbf{x}}, \tilde{\mathbf{y}})$, is the augmented data point and can be used as additional training data.

\xhdr{Proposed architecture change} Now, for the RE task we need to extract a fixed-length relation representation from the encoder output representation of the entire input sequence. The traditional way to do this for RE task is by concatenating the encoder output representations of the entity marker tokens \texttt{[E1]} and \texttt{[E2]}. However, it is challenging to do this for a virtual sample, $\tilde{\mathbf{x}}$, as the entity markers are not clearly defined in this case. We thus modify the relation representation to be the encoder output representation of the \texttt{[CLS]} token.
However, \citet{mtb} have shown this choice to be sub-optimal compared to concatenation of marker tokens. 
This is because the marker token representations provide direct access to the contextual information of the respective entities. Although the \texttt{[CLS]} token, in theory, has access to the entire context of the sentence, it might be difficult to capture the nuances like the head entity type, tail entity type, and the contextual information around the two entities all in a single vector. 

On the other hand, entity type information is easily accessible in most RE benchmarks\footnote{From new datasets/applications viewpoint, when entities are identified in a piece of text it is safe to assume that their types would also be identified.}.
So, to compensate for the sub-optimal choice of using \texttt{[CLS]} token representation as the relation representation, we modify how we represent the entity spans in the input token to more effectively use the easily accessible entity type information.
In particular, we note that the entity type labels can trivially be mapped to tokens from any pre-trained language model's vocabulary.
For example, entity types like \texttt{PERSON} and \texttt{STATE\_OR\_PROVINCE} can be tokenized into a word/phrase like `person' and `state or province', respectively. In such cases when entity type information is available, instead of using special marker tokens like \texttt{[E1]} and \texttt{[E2]} we prepend the entity spans in the input sequence with the word/phrases corresponding to their respective types and enclose these `type-words' in punctuation marks \cite{re_new_baseline}. The modified input to the transformer along with the \texttt{[CLS]} token looks as follows:
\begin{multline*}
\smallsize
    \texttt{[CLS]} \text{@ * person * \textbf{Lebron James} @ plays} \\
    \text{for \& * organization * \textbf{LA Lakers} \& team.}
\end{multline*}
We use different punctuation symbols to distinguish between subject and object entities. Specifically, we use `@' for subject and `\&' for object entities. This representation helps leverage the knowledge already contained in the pre-trained large-language model about the type of the entity and offset some of the downside of using a simplified relation representation in the \texttt{[CLS]} token.
As we will empirically see in \S\ref{sec:ablation}, this use of entity type information is not only effective but is necessary for the optimal functioning of our approach.
\citet{re_new_baseline} recently showed the success of this method in the fully supervised setting. Here we use it in conjunction with a simplified relation representation and show its merit in semi-supervised RE setting.

\subsection{Consistency training for SSRE}
\label{sec:ssre}

Let the given limited labelled set be $\mathbf{X}_l = \{\mathbf{x}_1^l, ..., \mathbf{x}_n^l\}$, with their relation labels $\mathbf{Y}_l = \{\mathbf{y}_1^l, ..., \mathbf{y}_n^l\}$, where $\mathbf{y}_i^l \in \{0, 1\}^{|\mathcal{R} \cup \{\text{NA}\}|}$ is a one-hot vector and $\mathcal{R}$ is the set of pre-defined relations. Let $\mathbf{X}_u = \{\mathbf{x}_1^u, ..., \mathbf{x}_m^u\}$ be a large unlabelled set. The goal is to apply both the data augmentation techniques described above and train a model with consistency loss to effectively leverage unlabelled data along with the limited labelled data.

We largely adapt the semi-supervised training techniques introduced by \citet{mixtext}. 
For each $\mathbf{x}_i^u$ in the unlabelled set $\mathbf{X}_u$, we generate $K$ augmentations $\mathbf{x}_{i,k}^a, k \in \{1, 2,..., K\}$ using the constrained back translation technique with different intermediate languages\footnote{In our specific implementation K = 2; with German and Russian as intermediate languages.}. 
These augmentations make up the set $\mathbf{X}_a = \{\mathbf{x}_{i,k}^a\}$.
For a given unlabelled data point $\mathbf{x}_i^u$ and its $K$ augmentations $\mathbf{x}_{i,k}^a$ the label is given by the average of current model's predictions on all $K+1$ data points:
$$
\mathbf{y}_i^u = \frac{1}{K + 1} \left(p_\phi(\mathcal{F}_{\theta}(\mathbf{x}_i^u)) + \sum_{k = 1}^{K}p_\phi(\mathcal{F}_{\theta}(\mathbf{x}_{i,k}^a))\right),
$$
where $\mathbf{y}_i^u$ is a probability vector. This not only enforces the constraint that the model should make consistent predictions for different augmentations but also makes the predictions more robust by ensembling all the predictions.
We merge the unlabelled set and the augmented set into $\mathbf{X}_{\text{ua}} = \mathbf{X}_u \cup \mathbf{X}_a$ and the corresponding (pseudo-)labels are given by $\mathbf{Y}_{\text{ua}} = \mathbf{Y}_u \cup \mathbf{Y}_a$, where $ \mathbf{Y}_u = \{\mathbf{y}_i^u\}$, $\mathbf{Y}_a = \{\mathbf{y}_{i,k}^a\}$, and $\mathbf{y}_{i,k}^a = \mathbf{y}_{i}^u \forall k \in \{1, 2,..., K\}$, i.e., all the augmented data points share the same label as the original unlabelled data point. 

Given this cumulative set $\mathbf{X}_{\text{ua}}$ and their generated labels $\mathbf{Y}_{\text{ua}}$ as additional training data, we employ the \name\ augmentation technique to generate arbitrary amounts of training data. In particular, we randomly sample two data points $\mathbf{x}_s^{\text{ua}}, \mathbf{x}_t^{\text{ua}} \in \mathbf{X}_{\text{ua}}$, and compute the encoder output representation of a new \textit{virtual} data point with $\text{REMix}(\mathbf{x}_s^{\text{ua}}, \mathbf{x}_t^{\text{ua}})$ and the corresponding target label with $\text{mix}(\mathbf{y}_s^{\text{ua}}, \mathbf{y}_t^{\text{ua}})$.

Additionally, while computing the final unsupervised loss in each training iteration we filter out the unlabelled data points with prediction confidence below a certain threshold $\gamma$ \cite{uda}. 
Finally, to encourage low-entropy predictions on unlabelled data, we sharpen the predictions with a sharpening coefficient $T$:
$$
\hat{\mathbf{y}}_i^{\text{ua}} = \frac{(\mathbf{y}_i^{\text{ua}})^{\frac{1}{T}}}{||(\mathbf{y}_i^{\text{ua}})^{\frac{1}{T}}||_1}.
$$
Everything put together, the final unsupervised loss in each training iteration with mini-batch size $B$ is computed as:
\begin{align*}
    \mathcal{L}_{\text{unsp}} = \frac{1}{B} \sum_{\mathbf{x}_s^{\text{ua}}, \mathbf{x}_t^{\text{ua}} \sim \mathbf{X}_{\text{ua}}}^B \texttt{m}(\mathbf{x}_s^{\text{ua}}, \mathbf{x}_t^{\text{ua}}) \mathcal{L}_{\text{mix}}(\mathbf{x}_s^{\text{ua}}, \mathbf{x}_t^{\text{ua}}),
\end{align*}
where 
\begin{align*}
\begin{split}
    \mathcal{L}_{\text{mix}}(\mathbf{x}_s^{\text{ua}}, \mathbf{x}_t^{\text{ua}}) ={}& \text{CE}(\text{mix}(\hat{\mathbf{y}}_s^{\text{ua}}, \hat{\mathbf{y}}_t^{\text{ua}}) || \\  & \qquad p_\phi(\text{REMix}(\mathbf{x}_s^{\text{ua}}, \mathbf{x}_t^{\text{ua}}))),
\end{split}\\
    \texttt{m}(\mathbf{x}_s^{\text{ua}}, \mathbf{x}_t^{\text{ua}}) ={}& I(\max\mathbf{y}_s^{\text{ua}} > \gamma)I(\max\mathbf{y}_t^{\text{ua}} > \gamma).
\end{align*}
Here, $I(.)$ is an indicator function and $\texttt{m}(.)$ denotes the confidence masking function which filters out the low-confidence datapoints. In our implementation $p_\phi(.)$ is a two-layer MLP classifier on top of the relation encoder model.
CE denotes the cross entropy loss function.
\footnote{Note that we only apply the augmentation techniques on the unlabelled data set. Initial experiments applying these to the labelled data set resulted in only marginal improvements and even performance deterioration in some cases, likely due to introduction of too much noise into an already limited labelled set.}

This combined with the traditional supervised loss, $\mathcal{L}_\text{sup} = \sum_{\mathbf{x}_i \sim \mathbf{X}_l}^{B} \text{CE}(\mathbf{y}_i^l||p_\phi(\mathcal{F}_{\theta}(\mathbf{x}_i)))$, constitutes the total loss:
$$
\mathcal{L}_{\text{total}} = \mathcal{L}_{\text{sup}} + \gamma_m \mathcal{L}_{\text{unsp}},
$$
where $\gamma_m$ is a parameter which controls the trade-off between supervised and unsupervised loss.
\section{Experiments}
We perform experiments on four benchmark datasets for sentence-level RE and compare the proposed model, \name, against current state-of-the-art SSRE approaches. We further conduct ablation studies and sensitivity analysis to expose the significance of different design choices of \name.
\subsection{Datasets}
We use two widely popular relation extraction benchmark datasets: SemEval 2010 Task 8 (SemEval) \cite{semeval}, and the TAC Relation Extraction Dataset (TACRED) \cite{tacred}. 
SemEval is a standard benchmark dataset for evaluating relation
extraction models containing 10717 examples in total. Each sentence is annotated with a pair of untyped nominals (concepts; example in Figure \ref{fig:re_aug}) that are related via one of 19 semantic relation types (including \texttt{no\_relation}). 
TACRED is a large-scale crowd-sourced relation extraction dataset with 106264 examples which is collected from all the prior TAC KBP relation schema. Unlike SemEval, sentences in TACRED are labelled with pairs of typed-entities that are related via one of 42 person- and organization-oriented relation types (including \texttt{no\_relation}).
In addition to these standard benchmark datasets, we also show results on two more datasets: RE-TACRED \cite{re-tacred} and KBP37 \cite{kbp37}.
RE-TACRED is a re-annotated version of the original TACRED dataset using an improved annotation strategy to ensure high-quality labels. \citet{re_new_baseline} provide a compelling analysis and recommend using this as the evaluation benchmark for sentence-level RE.
KBP37 is another sentence-level RE dataset with 21046 total examples collected from 2010 and 2013 KBP documents as well as July 2013 dump of Wikipedia. In terms of size, this falls between SemEval and TACRED. Similar to SemEval the entity types are not available in this dataset, however the 37 relation types are person- and organization-oriented like in TACRED. This dataset is thus a good segue between the two standard benchmarks.
The statistics of these datasets is given in Table \ref{tab:dataset_stats}.
\begin{table}[t]
    \caption{Dataset statistics}
    \centering
    \begin{tabular}{@{}llll@{}}
        \toprule
         Dataset & \# rel. & examples & \# no\_relation \\
        \midrule
        TACRED & 42 & 106264 & 79.51\%\\
        RE-TACRED & 40 & 91467 & 63.17\% \\
        KBP37 & 37 & 21046 & 10.33\% \\
        SemEval & 19 & 10717 & 17.39\% \\
        \bottomrule
    \end{tabular}
    \label{tab:dataset_stats}
\end{table}
\begin{table*}[th!]
    \caption{F1 score with various amounts of labelled data and 50\% unlabelled data. Mean and standard deviation of 5 different runs is reported. Best performance on each configuration is bolded and second best is underlined.}
    \centering
    \begin{tabular}{@{}lllllll@{}}
        \toprule
         & \multicolumn{3}{c}{TACRED} & \multicolumn{3}{c}{KBP37}\\
         \cmidrule(lr){2-4} \cmidrule(lr){5-7}
        \%labelled Data & 3\% & 10\% & 15\% & 5\% & 10\% & 30\% \\
        \midrule
        MRefG & 43.81\footnotesize $\pm$ 1.44 &	55.42\footnotesize $\pm$ 1.40 & 58.21\footnotesize $\pm$ 0.71 & - & - &	- \\
        MetaSRE & 46.16\footnotesize $\pm$ 0.74 &	56.95\footnotesize $\pm$ 0.33 & 58.94\footnotesize $\pm$ 0.31 &	59.29\footnotesize $\pm$ 0.92 &	61.83\footnotesize $\pm$ 0.21 &	63.51\footnotesize $\pm$ 0.69 \\
        GradLRE & \underline{47.37}\footnotesize $\pm$ 0.74 &	\underline{58.20}\footnotesize $\pm$ 0.33 & \underline{59.93}\footnotesize $\pm$ 0.31 &	\underline{59.98}\footnotesize $\pm$ 0.37 &	\underline{62.67}\footnotesize $\pm$ 0.54 &	\textbf{66.41}\footnotesize $\pm$ 0.28 \\
        REMix(ours) & \textbf{55.80}\footnotesize $\pm$ 1.33 &	\textbf{61.30}\footnotesize $\pm$ 0.70 & \textbf{63.07}\footnotesize $\pm$ 0.93 &	\textbf{60.84}\footnotesize $\pm$ 0.40 &	\textbf{63.82}\footnotesize $\pm$ 0.71 &	\textbf{66.46}\footnotesize $\pm$ 0.69 \\
        \midrule
        \midrule
        & \multicolumn{3}{c}{RE-TACRED} & \multicolumn{3}{c}{SemEval}\\
         \cmidrule(lr){2-4} \cmidrule(lr){5-7}
        \%labelled Data & 3\% & 10\% & 15\% & 5\% & 10\% & 30\% \\
        \midrule
        MRefG &	- &	- &	- &	75.48\footnotesize $\pm$ 1.34 &	77.96\footnotesize $\pm$ 0.90 & 83.24\footnotesize $\pm$ 0.71\\
        MetaSRE & 44.42\footnotesize $\pm$ 3.02 & 58.71\footnotesize $\pm$ 1.70 &	61.71\footnotesize $\pm$ 3.70 &	\underline{78.33}\footnotesize $\pm$ 0.92 &	80.09\footnotesize $\pm$ 0.78 & 84.81\footnotesize $\pm$ 0.44\\
        GradLRE & \underline{61.22}\footnotesize $\pm$ 0.58 & \underline{74.03}\footnotesize $\pm$ 1.74 &	\textbf{79.46}\footnotesize $\pm$ 0.82 &	\textbf{79.65}\footnotesize $\pm$ 0.68 &	\textbf{81.69}\footnotesize $\pm$ 0.57 & \textbf{85.52}\footnotesize $\pm$ 0.34\\
        REMix(ours) & \textbf{71.33}\footnotesize $\pm$ 1.22 & \textbf{77.94}\footnotesize $\pm$ 0.59 &	\textbf{79.76}\footnotesize $\pm$ 0.47 &	77.58\footnotesize $\pm$ 0.59 &	\underline{81.13}\footnotesize $\pm$ 0.82 & \textbf{85.51}\footnotesize $\pm$ 0.38 \\
        \bottomrule
    \end{tabular}
    \label{tab:main_results}
\end{table*}
\subsection{Baselines and implementation details}
We compare \name\ with three state-of-the-art models that are representative of the existing class of methods for SSRE: MRefG \cite{mrefg}, MetaSRE \cite{Hu2021SemisupervisedRE}, and GradLRE \cite{Hu2021GradientIR}. MRefG leverages the unlabelled data by semantically or lexically connecting them to labelled data by constructing reference graphs, such as entity reference or verb reference. This approach heavily leverages the linguistic structure of the data and is the only existing method that falls outside the \textit{self-training} class of methods. MetaSRE generates pseudo labels on unlabelled data by learning from the mistakes of the classification model as an additional meta-objective. GradLRE on the other hand generates pseudo label data to imitate the gradient descent direction on labelled data and bootstrap its optimization capability through trial and error \cite{Hu2021GradientIR}. MetaSRE and GradLRE are two of the strongest methods in the widely adapted \textit{self-training} methods for SSRE.

\xhdr{Implementation details} 
\todo{
\begin{itemize}
    \item add intermediate languages in BT for different datasets
\end{itemize}
}
We follow the established setting to use stratified sampling to divide the training set into various proportions of labelled and unlabelled sets so that the relation label distribution remains the same across all subsets. Following existing work, we sample 5\%, 10\%, and 30\% of the training set as labelled data for the SemEval and KBP37 datasets, and 3\%, 10\%, and 15\% of the training set as labelled data for TACRED and RE-TACRED datasets. For all datasets and experiments, unless otherwise specified, we sample 50\% of the training set as the unlabelled set.
For TACRED and SemEval datasets we take the performance numbers of all baseline models reported by \citet{Hu2021GradientIR}. For other datasets, we re-run the models with their best configuration as provided in their respective implementations, when available.
To be consistent with all the baselines we initialize the text encoder of \name\ with the \texttt{bert-base-cased} model architecture and pre-trained weights. 
Full details of all the hyperparameters can be found in Appendix \ref{sec:hyperparameters}.
\subsection{Main Results}
Table \ref{tab:main_results} shows F1 results of all baseline models and our proposed model, \name, on the four datasets when leveraging various amounts of labelled data and 50\% unlabelled data. We report the mean and standard deviation of 5 different runs (with different seeds) of training and testing.
\name\ gives state-of-the-art performance on 10 out of 12 different configurations across all four datasets.
This reinforces the importance of consistency regularization beyond the currently popular self-training methods for SSRE. 
Interestingly, the performance gains are significantly higher for TACRED and RE-TACRED datasets--we see an average improvement of as much as 17\% when trained on 3\% labelled data. 
This can be attributed to the fact that entity type information is available for these datasets and entity type markers are very effective in exploiting the knowledge embedded in the pre-trained language models.
We revisit this observation in our ablation studies (\S\ref{sec:ablation}) where we concretely establish the benefits of using entity type markers.
\subsection{Analysis and discussion}
\label{sec:ablation}
\begin{figure*}[h]
  \centering
  \centerline{\includegraphics[width=\linewidth]{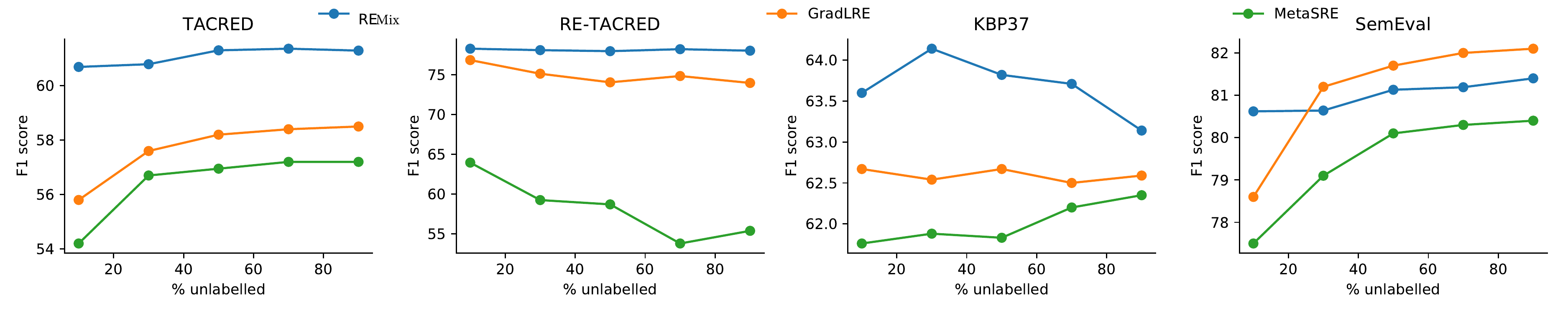}}
\caption{F1 Performance with various unlabelled data and 10\% labelled data}
\label{fig:ablation2}
\end{figure*}
\begin{figure*}
  \centering
  \centerline{\includegraphics[width=\linewidth]{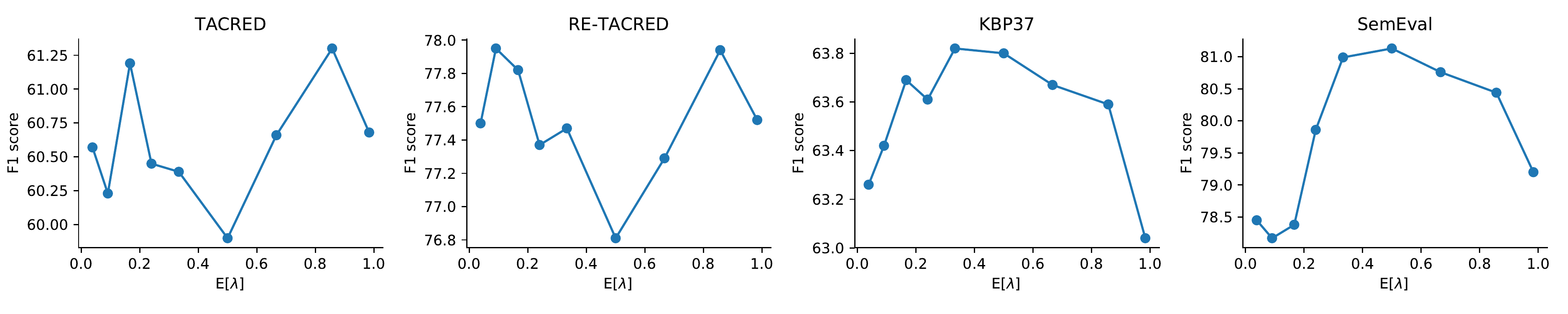}}
\caption{F1 Performance of \name\ with 50\% unlabelled and 10\% labelled data with changing mixing coefficient $\lambda$}
\label{fig:ablation3}
\end{figure*}
We first conduct experiments to empirically demonstrate the effectiveness of three components of our proposed model: i) data augmentation by latent space interpolation (Mix-DA), ii) data augmentation by constrained back-translation (BT-DA), and iii) entity type markers (ET).
In Table \ref{tab:ablation_results}, we report the mean F1 score of five different runs for different variations of our model by removing a certain combination of these components\footnote{We omit standard deviation values for brevity. See the full results and comparisons in Appendix \ref{sec:full_ablation}}.
As can be seen from Table \ref{tab:ablation_results}, each of these components contributes to the overall success of \name.
For contribution of just the Mix-DA: we compare i) row 1 v/s row 3, and ii) row 2 v/s row 4. All comparisons show positive improvement. 
For contribution of just the BT-DA: we compare i) row 1 v/s row 2, and ii) row 3 v/s row 4. We note that BT-DA results only in marginal improvements in most cases. 
Upon closer inspection we note that the constrained-decoding algorithms we implement for BT-DA are actually not perfect, especially when combined with translation models. It sometimes misses the constraints and sometimes falls into repetitive loops in an attempt to satisfy the constraint. With the ever-improving language generation capabilities, we believe the quality of data augmentation will only improve with time and result in more significant performance improvements.
For contribution of both DA techniques together: we compare row 1 v/s row 4. All comparisons show significant improvements with data augmentation.
The contribution of entity type markers can be noted in TACRED and RE-TACRED datasets. We see an average drop of 5.4\% in F1 score across all 8 comparisons. Although our data augmentation techniques are effective, without the entity type information our method doesn't result in state-of-the-art performance. This reinforces our architectural choice to include entity type markers when using \texttt{[CLS]} token for relation representation (\S\ref{sec:mixre}).
\begin{table}[th!]
    \caption{Ablation results on all datasets using 10\% labelled set and 50\% unlabelled set.}
    \centering
    \begin{tabular}{@{}lllllll@{}}
        \toprule
        \scriptsize Mix. & \scriptsize BT-DA & \scriptsize ET & \scriptsize TACRED & \scriptsize RE-T & \scriptsize $\text{KBP37}^*$ & \scriptsize $\text{SemEval}^*$\\
        \midrule
        \multirow{2}{*}{\cmark} & \multirow{2}{*}{\cmark} & \cmark & 
        \small61.30 &	
        \small77.94 & 
        \multirow{2}{*}{\small63.82} & 
        \multirow{2}{*}{\small81.13} \\
         &  & \xmark & 
         \small56.82 &	
         \small75.11 & 
          & 
          \\
        \midrule
        \multirow{2}{*}{\cmark} & \multirow{2}{*}{\xmark} & \cmark & 
        \small60.81 &	
        \small77.77 & 
        \multirow{2}{*}{\small63.48} & 
        \multirow{2}{*}{\small79.71} \\
         &  & \xmark & 
         \small56.35 &	
         \small74.67 & 
          & 
          \\
        \midrule
        \midrule
        \multirow{2}{*}{\xmark} & \multirow{2}{*}{\cmark} & \cmark & 
        \small59.65 &	
        \small76.80 & 
        \multirow{2}{*}{\small62.64} & 
        \multirow{2}{*}{\small79.17} \\
         &  & \xmark & 
         \small55.52 &	
         \small73.78 & 
          & 
          \\
        \midrule
        \multirow{2}{*}{\xmark} & \multirow{2}{*}{\xmark} & \cmark & 
        \small58.96 &	
        \small77.25 & 
        \multirow{2}{*}{\small63.14} & 
        \multirow{2}{*}{\small79.20} \\
         &  & \xmark & 
         \small55.25 &	
         \small74.58 & 
          & 
          \\
        \bottomrule
    \end{tabular}
    \begin{flushleft}
    \footnotesize * these datasets do not have entity type information
    \end{flushleft}
    \label{tab:ablation_results}
\end{table}

Next we examine the effect of using different amounts of unlabelled data. In Figure \ref{fig:ablation2}, we report the average F1 score for different models trained with different amounts of unlabelled data and 10\% labelled data. \name\ outperforms the baselines in all settings except on SemEval dataset, and, interestingly, the performance only marginally changes with the change in the amount of unlabelled data.
Note that we train the models until the performance on the validation set stops improving for more than 5 epochs.
Hence, \name\ generates, in principle, an infinite amount of unlabelled data via the mixup strategy. Coupled with the fact that the label distribution remains the same in all settings, adding more unlabelled data does not seem to add a lot of new information. 
This explains why the model performance is relatively insensitive to changing amounts of unlabelled data.
This also implies that \name\ can leverage low amounts of unlabelled data better than the baselines.

Finally, in Figure \ref{fig:ablation3} we show how the performance of \name\ changes with a change in the mean of the Beta distribution from which $\lambda$ is sampled on each iteration.
Note that a value near 0 and 1 for $\lambda$ means the augmented \textit{virtual} data point will be closer to one of the underlying data points. As we get closer to 0.5 the virtual data points get further from the original data manifold and become more `novel'.
On TACRED and RE-TACRED datasets the performance peaks at $\text{E}(\lambda) = 0.15 (\text{or 0.85})$ and drops in the mid-values. 
This can be interpreted as: adding datapoints far from the original data manifold is detrimental for these datasets. 
Interestingly, on KBP37 and SemEval the pattern inverts, i.e., the performance increases as $\text{E}(\lambda)$ approaches $0.5$, implying that more `novel' augmentations help for these datasets.

\section{Conclusion}
In this paper, we propose a consistency-training-based semi-supervised algorithm for relation extraction and empirically show the merit of this class of methods in comparison to the current state-of-the-art \textit{self-training} class of methods.
In future work, one could bootstrap the self-training methods with consistency training as done in some previous works on vision tasks \cite{Pham2021MetaPL}.
Additionally, we show how the entity type information, when available, can result in massive performance boosts in the semi-supervised scenario. This is important because in most practical use cases when entities have already been identified, the entity type information is easy available and could be effectively leveraged in the proposed fashion.

\section{Limitations}
One of the key limitations of our proposed method compared to the baseline methods is the tight dependence on a strong external translation system to get good quality back-translated data augmentations.
Secondly, since we use \texttt{[CLS]} token embeddings instead of entity-specific embeddings for final classification, it is more challenging to decipher entity-specific context. This is evident from the fact that our method performs relatively the weakest on the SemEval dataset which consists of untyped nominals (concepts) as entities and abstract relations which we believe need more entity-specific context to understand. Hence, our proposed method, \name, shines bright when the entities in the dataset are typed or named entities whose meaning or type is relatively context-agnostic.

\section*{Acknowledgements} The author would like to thank the reviewers for their thoughtful feedback, the entire AI/ML Research team at Vanguard for their support, and especially David Landsman for their valuable input and discussion on the initial drafts of this work.

\section*{Disclaimer} This material is provided for informational purposes only and is not intended to be investment advice or a recommendation to take any particular investment action.
\textcopyright 2022 The Vanguard Group, Inc. All rights reserved.

\bibliography{anthology,custom}
\bibliographystyle{acl_natbib}

\clearpage
\appendix

\section{Reproducibility checklist}
\label{sec:reproducibility}
\subsection{Datasets} The sources of all the datasets are given in Table \ref{tab:data_sources}. We use the given train/validation/test splits for TACRED, RE-TACRED and KBP37 datasets. For SemEval dataset, we use the same splits as all the baselines, i.e., we split the original training set into 90\% training set and 10\% validation set.

\begin{table*}[b]
    \caption{Dataset sources}
    \centering
    \begin{tabular}{@{}ll@{}}
        \toprule
         Dataset & Source \\
        \midrule
        TACRED & \url{https://catalog.ldc.upenn.edu/LDC2018T24} \\
        RE-TACRED & \url{https://github.com/gstoica27/Re-TACRED} \\
        KBP37 & \url{https://github.com/zhangdongxu/kbp37} \\
        SemEval & \url{https://semeval2.fbk.eu/semeval2.php?location=data} \\
        \bottomrule
    \end{tabular}
    \label{tab:data_sources}
\end{table*}

\subsection{Hyperparameters} 
\label{sec:hyperparameters}
We use the BERT tokenizer and set maximum sequence length to 256 to pre-process all datasets. 
We use the AdamW optimizer \cite{adamw} with 5e-5 learning rate and 0.1 warmup ratio. 
We sweep over the following hyperparameters: sharpening coefficient $T$, confidence threshold $\gamma$, the Beta-distribution parameters ($\alpha$, $\beta$)\footnote{$\alpha$ is fixed to be 60 and we change the values of $\beta$ to control the mean of the distribution}, and the unsupervised loss weight $\gamma_m$. 
We perform incremental grid search to get the best performing configuration based on the F1 score on validation set. Table \ref{tab:hyperparameter_grid} shows the set of values we use for each parameter. Table \ref{tab:best_config} shows the best parameter values on each dataset and configuration.
Following \citet{mixtext}, we use \{7, 9, 12\} for the mixup layer set; this layer subset contains most of the syntactic and semantic information as suggested by \citet{jawahar-etal-2019-bert}.
    
\begin{table}[H]
    \caption{Hyperparameter search values}
    \centering
    \begin{tabular}{@{}ll@{}}
        \toprule
        Parameter & Values \\
        \midrule
        $T$ & \{0.4, 0.6, 0.8, 1.0\} \\
        $\gamma$ & \{0, 0.15, 0.2, 0.25\} \\
        $\beta$ & \{1, 10, 30, 60, 120, 190, 300, 600\}* \\
        $\gamma_m$ & \{0.01, 0.1, 1\} \\
        \bottomrule
    \end{tabular}
    \begin{flushleft}
    * corresponding means of the sampled mixing coefficient, $\lambda$, are given by \{0.04, 0.09, 0.17, 0.24, 0.33, 0.50, 0.67, 0.86, 0.98\}
    \end{flushleft}
    \label{tab:hyperparameter_grid}
\end{table}

\begin{table}[H]
    \caption{Best hyperparameter values}
    \centering
    \begin{tabular}{@{}lllllll@{}}
        \toprule
         & \multicolumn{3}{c}{TACRED} & \multicolumn{3}{c}{KBP37}\\
         \cmidrule(lr){2-4} \cmidrule(lr){5-7}
         & 3\% & 10\% & 15\% & 5\% & 10\% & 30\% \\
        \midrule
        $T$ & 0.8 &	0.8 & 0.8 & 0.8 & 0.8 & 0.8 \\
        $\gamma$ & 0.15 &	0.15 & 0.90 & 0.25 &	0.25 & 0.70	\\
        $\beta$ & 10 & 10 & 10 & 120 & 120 & 190 \\
        $\gamma_m$ & 0.01 & 0.1 & 0.1 & 0.1 & 1.0 & 1.0 \\
        \midrule
        \midrule
        & \multicolumn{3}{c}{RE-TACRED} & \multicolumn{3}{c}{SemEval} \\
         \cmidrule(lr){2-4} \cmidrule(lr){5-7}
         & 3\% & 10\% & 15\% & 5\% & 10\% & 30\% \\
        \midrule
        $T$ & 0.4 &	0.4 & 0.4 & 0.8 & 0.8 & 0.8 \\
        $\gamma$ & 0.0 &	0.0 & 0.9 & 0.2 &	0.2 & 0.7	\\
        $\beta$ & 10 & 10 & 10 & 60 & 60 & 60 \\
        $\gamma_m$ & 0.01 & 0.1 & 0.1 & 0.1 & 1.0 & 1.0 \\
        \bottomrule
    \end{tabular}
    \label{tab:best_config}
\end{table}

\subsection{Training details} We train each model on a single NVIDIA Tesla T4 GPU with 16GB memory. We employ mixed precision training and gradient checkpointing techniques for faster and memory-efficient training. 
Note that we train the models until the performance on the validation set plateaus. 
The full \name\ model roughly takes about 6 hours to train on TACRED, 5 hours in RE-TACRED, 1 hour in KBP37, and about 30 minutes on SemEval. Note that the training time slightly varies ($\pm$ 30 minutes) depending on the percentage of labelled and unlabelled data we use. The number of parameters in all our models are largely dominated by the \texttt{bert-base-cased} that we use as the text encoder. The relatively negligible varying component is the MLP classifier that varies with the varying number of relations in each dataset.

\begin{table*}
    \caption{F1 score with 10\% of labelled data and 50\% unlabelled data. Mean and standard deviation of 5 different runs is reported.}
    \centering
    \begin{tabular}{@{}llllllll@{}}
        \toprule
         & Mix. & BT aug. & Ent. type & TACRED & RE-TACRED & KBP37 & SemEval\\
        \midrule
        \multirow{2}{*}{a)} & \multirow{2}{*}{\cmark} & \multirow{2}{*}{\cmark} & \cmark & 
        61.30\footnotesize $\pm$ 0.70 &	
        77.94\footnotesize $\pm$ 0.59 & 
        \multirow{2}{*}{63.82\footnotesize $\pm$ 0.71} & 
        \multirow{2}{*}{81.13\footnotesize $\pm$ 0.82} \\
         & &  & \xmark & 
         56.82\footnotesize $\pm$ 0.64 &	
         75.11\footnotesize $\pm$ 1.16 & 
          & 
          \\
        \midrule
        \multirow{2}{*}{b)} & \multirow{2}{*}{\cmark} & \multirow{2}{*}{\xmark} & \cmark & 
        60.81\footnotesize $\pm$ 1.31 &	
        77.77\footnotesize $\pm$ 0.96 & 
        \multirow{2}{*}{63.48\footnotesize $\pm$ 0.53} & 
        \multirow{2}{*}{79.71\footnotesize $\pm$ 0.83} \\
         & &  & \xmark & 
         56.35\footnotesize $\pm$ 0.97 &	
         74.67\footnotesize $\pm$ 1.04 & 
          & 
          \\
        \midrule
        \midrule
        \multirow{2}{*}{c)} & \multirow{2}{*}{\xmark} & \multirow{2}{*}{\cmark} & \cmark & 
        59.65\footnotesize $\pm$ 0.92 &	
        76.80\footnotesize $\pm$ 0.98 & 
        \multirow{2}{*}{62.64\footnotesize $\pm$ 0.69} & 
        \multirow{2}{*}{79.17\footnotesize $\pm$ 1.64} \\
         & &  & \xmark & 
         55.52\footnotesize $\pm$ 0.89 &	
         73.78\footnotesize $\pm$ 1.34 & 
          & 
          \\
        \midrule
        \multirow{2}{*}{d)} & \multirow{2}{*}{\xmark} & \multirow{2}{*}{\xmark} & \cmark & 
        58.96\footnotesize $\pm$ 1.21 &	
        77.25\footnotesize $\pm$ 0.70 & 
        \multirow{2}{*}{63.14\footnotesize $\pm$ 0.90} & 
        \multirow{2}{*}{79.20\footnotesize $\pm$ 0.32} \\
         & &  & \xmark & 
         55.25\footnotesize $\pm$ 1.53 &	
         74.58\footnotesize $\pm$ 0.91 & 
          & 
          \\
        \bottomrule
    \end{tabular}
    \label{tab:full_ablation_results}
\end{table*}

\section{Full Ablation results}
\label{sec:full_ablation}
We conduct experiments to empirically demonstrate the effectiveness of three components of our proposed model: i) data augmentation by latent space interpolation, ii) data augmentation by constrained back-translation, and iii) entity type markers.
In Table \ref{tab:full_ablation_results}, we report the performance of different variations of our model by removing a certain combination of these components.
As can be seen from Table \ref{tab:full_ablation_results}, each of these components contributes to the overall success of \name. 
To see the impact of data augmentation by latent space interpolation, we compare the results in row `a' with the corresponding ones in row `c', and similarly row `b' with row `d'. 
We see a significant and consistent drop in every comparison with and without that component. Specifically, we see a drop of an average 1.6\% in F1 score over all the 12 comparisons.
Another interesting pattern that stands out is the significant and consistent drop in performance (average 5.4\% in F1 score across all 8 comparisons) when not using entity type markers.
When we drop the entity type markers from the input representation, we use the basic entity start and end markers--\texttt{[E1],} \texttt{[/E1]}, \texttt{[E2],} \texttt{[/E2]}. 
This only applies to TACRED and RE-TACRED since only these datasets have entity type information readily available.
Note that when we do not use any of the data augmentations, consistency training is not possible and we are effectively using only the labelled set for training the model (fourth row in Table \ref{tab:full_ablation_results}).
Surprisingly, even without any unlabelled data using the entity type markers alone gives better performance than the current state-of-the-art results by GradLRE \cite{Hu2021GradientIR}, i.e., an F1 score of 58.96 vs 58.20 on TACRED and 77.25 vs 74.03 on RE-TACRED. 
This proves the effectiveness of using entity type information in the fashion we use, i.e., by tokenizing the type words to leverage the knowledge embedded in pre-trained language models.

\end{document}